\documentclass{article}

\PassOptionsToPackage{numbers, compress}{natbib}
\usepackage[preprint]{neurips_2023}
\usepackage[utf8]{inputenc} % allow utf-8 input
\usepackage[T1]{fontenc}    % use 8-bit T1 fonts
\usepackage{hyperref}       % hyperlinks
\usepackage{url}            % simple URL typesetting
\usepackage{booktabs}       % professional-quality tables
\usepackage{amsfonts}       % blackboard math symbols
\usepackage{nicefrac}       % compact symbols for 1/2, etc.
\usepackage{microtype}      % microtypography
\usepackage{xcolor}         % colors
\usepackage{multirow}
\usepackage{graphicx}
\usepackage{amsmath}
\usepackage{cleveref}
\crefname{section}{Sec.}{Secs.}
\Crefname{section}{Section}{Sections}
\Crefname{table}{Table}{Tables}
\crefname{table}{Tab.}{Tabs.}
\crefname{figure}{Fig.}{Figs.}
\crefname{algorithm}{Alg.}{Algs.}
\crefformat{equation}{Eq.~#2#1#3}

\newcommand{\eg}{\textit{e}.\textit{g}.}

\newcommand{\ie}{\textit{i}.\textit{e}.}
\newcommand{\wrt}{\textit{w}.\textit{r}.\textit{t}.}
\newtheorem{theorem}{Theorem}
\newtheorem{definition}{Definition}

\definecolor{hollywoodcerise}{rgb}{0.96, 0.0, 0.63}
\definecolor{lasallegreen}{rgb}{0.03, 0.47, 0.19}
\definecolor{hanpurple}{rgb}{0.32, 0.09, 0.98}
\definecolor{green(pigment)}{rgb}{0.0, 0.65, 0.31}
\usepackage{hyperref}
\hypersetup{colorlinks,linkcolor={red},citecolor={hollywoodcerise},urlcolor={red}}  

\title{FMapping: Factorized Efficient Neural Field Mapping for Real-Time Dense RGB SLAM}

\author{%
  Tongyan Hua$^{\dagger}$ \\
  AI Thrust, HKUST(GZ)\\
  \texttt{tongyanhua@hkust-gz.edu.cn} \\
   \And
   Haotian Bai$^{\dagger}$\\
  AI Thrust, HKUST(GZ)\\
   \texttt{haotianwhite@outlook.com} \\
   \AND
   Zidong Cao \\
  AI Thrust, HKUST(GZ)\\
   \texttt{caozidong1996@gmail.com} \\
   \And
   Lin Wang\thanks{Corresponding author. $^{ \dagger}$ Authors with equal contribution.} \\
  AI Thrust, HKUST(GZ)\\
  Dept. of Computer Science and Engineering, HKUST\\
   \texttt{linwang@ust.hk} \\
}

\begin{document}

\maketitle

\begin{abstract}

%Background
% Simultaneous Localization and Mapping(SLAM) involves constructing a map of an unknown environment while simultaneously estimating the robot's current location, which is typically divided into initialization, tracking, and mapping. During initialization, the system establishes an initial state, while in the tracking stage, it estimates poses for mapping to update the reconstructed scene continuously.

% % What have other people done about this problem? 

% Previous studies have attempted to reduce the complexity of implicit dense mapping by utilizing geometric priors from depth sensors, pre-trained models, or explicit point cloud maps from external SLAM systems. 
In this paper, we introduce \textbf{FMapping}, an efficient neural field mapping framework that facilitates the continuous estimation of a colorized point cloud map in real-time dense RGB SLAM. To achieve this challenging goal without depth, a hurdle is how to improve efficiency and reduce the mapping uncertainty of the RGB SLAM system. 
To this end, we first build up a theoretical analysis by decomposing the SLAM system into tracking and mapping parts, and the mapping uncertainty is explicitly defined within the frame of neural representations. Based on the analysis, we then propose an effective factorization scheme for scene representation and introduce a sliding window strategy to reduce the uncertainty for scene reconstruction. Specifically, we leverage the factorized neural field to decompose uncertainty into a lower-dimensional space, 
%\ie, a 4D continuous radiance field, 
which enhances robustness to noise and improves training efficiency. 
We then propose the sliding window sampler to reduce uncertainty  by incorporating coherent geometric cues from observed frames during map initialization to enhance convergence. 
Our factorized neural mapping approach enjoys some advantages, such as low memory consumption, more efficient computation, and fast convergence during map initialization.
Experiments on two benchmark datasets show that our method can update the map of high-fidelity colorized point clouds around 2 seconds in real time while requiring no customized CUDA kernels. Additionally, it utilizes $\times 20$ fewer parameters than the most concise neural implicit mapping of prior methods for SLAM, \eg, iMAP~\cite{DBLP:conf/iccv/SucarLOD21} and around $\times 1000$ fewer parameters than the state-of-the-art approach, \eg, NICE-SLAM~\cite{DBLP:conf/cvpr/ZhuPLXBCOP22}. For more details, please refer to our project homepage: \url{https://vlis2022.github.io/fmap/}.

\end{abstract}

\vspace{-20pt}
\section{Introduction}
\vspace{-5pt}
%\begin{figure}[t!]
%    \centering
%    \includegraphics[width=0.42\textwidth]{images/Teaser_figure.pdf}
%        \vspace{-12pt}
%    \caption{(a) The changing trend regarding the comparison of the total number of papers with that using deep learning for event-based vision. (b) An overview of our survey methodology.
    % about deep learning for event-based vision.
%    }
%    \label{fig:Comparison}
%    \vspace{-15pt}
%\end{figure}

% What is Dense RGB SLAM & why interest

% 要有铺垫，多描述清楚SLAM是啥
% Visual SLAM is a great domain that been addressed the
Dense visual Simultaneous Localization and Mapping (SLAM) aims to build a map of an unknown environment while simultaneously estimating the camera pose from the inputs. It is typically divided into initialization, tracking, and mapping. During the initialization, the system establishes an initial state, while in the tracking stage, it estimates poses for the mapping stage to update the reconstructed scene continuously.
Earlier methods,~\eg,~\cite{henry2014rgb, dai2017bundlefusion} are often built based on the RGB-D cameras. However, these methods are not applicable when no depth sensors are available, as they have difficulty in estimating accurate geometric cues.
% However, depthcameras are limited to indoor environments and are more expensive to obtain. 

% distant, reflective, and transparent regions, which cause great difficulties in their applications. 
% However, RGB-D cameras usually suffer from limited range and are sensitive to lighting conditions, reflective surfaces, or occlusions.

Therefore, it is valuable to explore dense RGB SLAM~\textemdash reconstructing a dense 3D map of the scene in real time from only regular RGB inputs. Such a SLAM paradigm offers notable advantages by eliminating the need for depth sensors and providing a comprehensive scene representation, beneficial for onboard tasks, such as robot navigation~\cite{temeltas2008slam,fang2021visual} and interactive digital applications~\cite{bettens2020unrealnavigation, sato2020construction}.
%
% challenges of mapping problems in Dense RGB SLAM
However, mapping in dense RGB SLAM brings unique difficulties, which can be categorized into two primary aspects.
% compared with the mapping in SLAM systems using external sensors,~\eg, LiDAR. These obstacles 
% can be categorized into two primary aspects. 
Firstly, explicitly storing large amounts of point cloud data 
% to process RGB images 
imposes a high requirement for computation and memory, limiting the practical application to resource-restricted mobile devices. 
% which is particularly critical when 
% considering that dense RGB SLAM systems are often deployed on the mobile devices.
Secondly, the lack of depth inputs leads to inherently noisy pixel-level depth estimation results, if solely relying on the cross-frame consistency.
% with no depth inputs, solely relying on cross-frame consistency leads to noisy depth estimation results.

%and Fig. \ref{fig:Comparison2}. 
%RGB-D solutions like NICE-SLAM~\cite{DBLP:conf/cvpr/ZhuPLXBCOP22}, iMAP~\cite{DBLP:conf/iccv/SucarLOD21} are unable to render the map in short optimization interval with camera on the fly, which highlights the mapping efficiency given relatively accurate camera positions. 
%
% How NeRF address these challenge
Neural Radiance Fields (NeRFs)~\cite{mildenhall2020nerf} have recently emerged as a compelling solution to the mapping problem for offline 3D reconstruction. NeRF utilizes latent representations, such as multi-layer perceptron (MLP) to implicitly estimate the density and color of 3D points. It transforms the explicitly-cached point clouds into a compact implicit representation, thereby reducing memory consumption.
Moreover, the physical model underlying NeRF enables rendering a 3D point based on its pixel-aligned density, allowing the gradient from color supervision to be extended to geometric estimation.
%
% Moreover, thanks to the physical model behind NeRF, a 3D point is rendered based on its pixel-aligned spatial occupancy estimation, which enables the gradient from color supervision to be extended to geometric estimation as well.
%
%
\begin{figure}[t]
\centering
\includegraphics[width=1\textwidth]{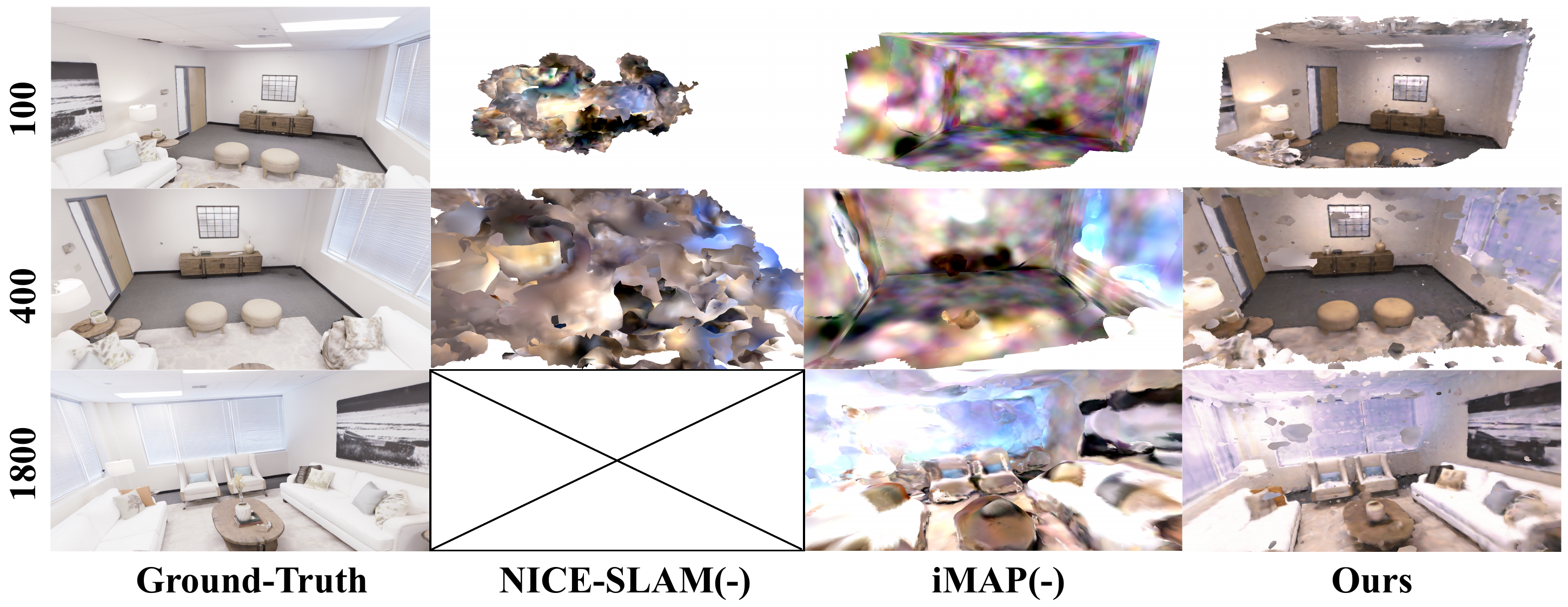}
\vspace{-15pt}
\caption{Dense mapping snapshots (at 100, 400, and 1800 input frames) of the on-the-fly running of NICE-SLAM(-)~\cite{DBLP:conf/cvpr/ZhuPLXBCOP22}, iMAP(-)~\cite{DBLP:conf/iccv/SucarLOD21}, and ours in full-length Replica (Room0) sequence, given ground truth (GT) poses without depth supervision. (-) denotes that we make modifications to the original implementations by eliminating the back-propagation of the gradient from depth supervision. The blank space in the 1800 frame of NICE-SLAM(-)~\cite{DBLP:conf/cvpr/ZhuPLXBCOP22} denotes running fails, and thus is removed from our experimental settings. In comparison, our method can reconstruct high-fidelity dense RGB maps throughout the entire SLAM process.}
\vspace{-16pt}
\label{fig:Comparison}
\end{figure}
%
%
% from scratch
% The geometric estimation is further stabilized and extended to larger scenes by NICE-SLAM~\cite{DBLP:conf/cvpr/ZhuPLXBCOP22}, which proposes to include hierarchical voxel grids and pre-trained geometric decoders. 
% Nevertheless, these works all strongly depend on depth supervision to constrain the uncertainty for on-the-fly updating of sample point distribution along the camera rays, given elusive camera poses. 
%
% mention of "NeRF RGB SLAM" and What are their cons compared to us.
% Efforts have been made by many recent NeRF-based methods to address the mapping problem in dense RGB SLAM.
%[\textbf{Note: +figure 1 description about iMAP and NICE-SLAM, their drawbacks}]
%
%Specifically, iMAP~\cite{DBLP:conf/iccv/SucarLOD21} employs only a single MLP to represent the entire scene efficiently, but is restricted to small scenes. NICE-SLAM improves the neural implicit representation with hierarchical voxel grids and pre-trained models, which can be generalized to large scenes. 
%
%The mapping uncertainty in iMAP and NICE-SLAM is decreased with geometric priors, either from depth sensors or geometric cues from pre-trained models.
%
Inspired by NeRF, some works, \eg, iMAP~\cite{DBLP:conf/iccv/SucarLOD21} and NICE-SLAM~\cite{DBLP:conf/cvpr/ZhuPLXBCOP22} propose to build neural implicit representations for dense RGB-D SLAM. These methods show that the mapping uncertainty can be decreased by introducing geometric priors, either from depth sensors or geometric cues from pre-trained models. 
Some recent NeRF-based methods~\cite{rosinol2022nerf,Li2023DenseRS,DBLP:journals/corr/abs-2302-03594,chung2022orbeez} have made similar efforts to tackle the mapping problem for dense RGB SLAM without depth. 
For instance, NeRF-SLAM ~\cite{rosinol2022nerf} incorporates geometric cues, \eg, point clouds, derived from external SLAM systems~\cite{teed2021droid}. DIM-SLAM ~\cite{Li2023DenseRS} performs multi-scale grid occupancy estimation using a cross-frame photometric warping loss. In a nutshell, these methods require extensive computation costs to obtain extra geometric cues in order to reach a similar level compared to the RGB-D SLAM system and are not readily applicable in real time without a customized CUDA implementation. In this paper, we observe that, \textit{even without explicit geometric cues or priors}, a dense RGB SLAM system indeed needs to \textit{reduce the mapping uncertainty} upon integration of NeRF, while \textit{improving the mapping efficiency}.

In light of this, we present \textbf{FMapping}, an efficient neural field mapping framework that facilitates the continuous estimation of a colorized point cloud map in real-time dense RGB SLAM.
% [external]
% To solve this issue while constraining the mapping uncertainty, 
To achieve this goal without depth, an obstacle is how to improve efficiency and reduce the implicit mapping uncertainty for RGB SLAM, and no theoretical support is available to examine the mapping uncertainty upon the integration of NeRF.
Therefore, we first conduct a theoretical analysis to model and decompose the uncertainty and allocate it to different variables (\cref{sec: Methodology}), which reveals the solution lies in the suitable map representation and sampling strategy. Also, we reformulate the mapping problem to maximize the posterior probability conditioned on the learned latent features, given incrementally observed sequences of RGB inputs. 
Based on the analysis, we then propose an effective factorization scheme for scene representation and 
introduce a sliding window strategy to reduce the uncertainty for scene reconstruction.
Specifically, 
% we formulate the mapping representation as a continuous latent radiance field with uniform distributed voxel grids, as inspired by the formulation in NICE-SLAM~\cite{DBLP:conf/cvpr/ZhuPLXBCOP22}.
% From another perspective, we explicitly model the scene as a 4D tensor, meaning that given a 3D point, we assign a voxel latent feature to represent its occupancy and entailed radiance.
inspired by TensoRF~\cite{Chen2022TensoRFTR}, we leverage the factorized neural field to decompose uncertainty into a lower-dimensional space, \ie, a 4D continuous radiance field (\cref{sec: f-map}). That is, we explicitly model the scene as a 4D tensor, where each 3D grid is associated with a voxel latent feature that represents its density and color.
Moreover, we propose the sliding window sampler to capture coherent geometric cues to handle excessive uncertainty in the initialization stage, which speeds up the convergence and provides more accurate pose estimation (\cref{sec:sws}).
%
% In this way, we reformulate the mapping problem in Dense RGB SLAM as a maximal posterior distribution estimation of the latent features, given incomplete sequences of RGB inputs. 
% To make this problem resolvable in real-time, \ie, faster convergency and efficient computation, our contribution is two-fold in a consecutive manner:
Consequently, our solution enables real-time map reconstruction involving efficient computation, low memory consumption, and fast convergence for dense RGB SLAM.

We show that our FMapping can reconstruct a high-fidelity dense map more efficiently for real-time RGB SLAM than existing methods, given the same good pose estimation in the on-the-fly stage, as shown in~\cref{fig:Comparison}. Note that our approach achieves efficiency and accuracy gains through a standard PyTorch implementation, in contrast to some NeRF-based SLAM methods~\cite{rosinol2022nerf,Li2023DenseRS} that rely on customized CUDA kernels. In summary,
our major contributions are three-fold: 
(\textbf{I}) We provide a theoretical analysis of the mapping uncertainty upon the integration of NeRF for dense RGB SLAM. 
(\textbf{II}) We introduce the factorized neural field to decompose uncertainty into a lower-dimensional space, and a sliding window sampling strategy to further reduce uncertainty by coherent geometric cues. 
(\textbf{III}) Our method achieves real-time high-fidelity map updating with a standard PyTorch implementation while utilizing around $\times 1000$ fewer parameters than the state-of-the-art approach~\cite{DBLP:conf/cvpr/ZhuPLXBCOP22,Li2023DenseRS}.

\vspace{-8pt}
\section{Related Work}
\vspace{-8pt}
\noindent \textbf{Dense Visual SLAM.} 	Dense visual SLAM has experienced rapid evolution in the past two decades. Compared to sparse visual SLAM algorithms~\cite{DBLP:conf/ismar/KleinM07,DBLP:journals/trob/Mur-ArtalT17} that reconstruct sparse point clouds, dense visual SLAM algorithms~\cite{DBLP:conf/iccv/NewcombeLD11} are able to recover dense point cloud representations of the scene. Some iconic traditional dense SLAM works~\cite{newcombe2011kinectfusion, keller2013real} explicitly represent surfaces using standard volume representation. In addition, some works~\cite{dai2017bundlefusion, vespa2018efficient} 
employ hierarchical volume representations, which offer increased efficiency but present challenges in implementation and parameter optimization due to their size. Recently, deep learning-based works~\cite{czarnowski2020deepfactors, li2020deepslam, li2018undeepvo} have made great advances in the dense visual SLAM, bringing the benefits of both accuracy improvement and robustness enhancement. \textit{Different from the aforementioned explicit representation methods, we focus on implicitly representing the scene, which is compact and can be extended to unobserved regions.}

\begin{figure}[t]
\centering
\includegraphics[width=1\textwidth]{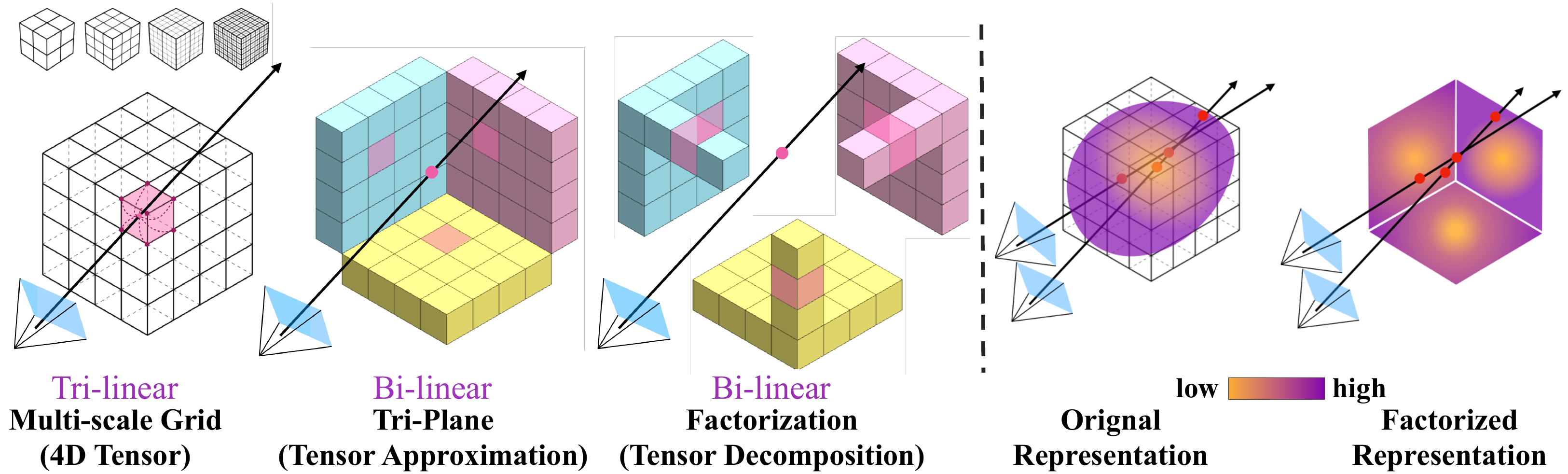}
\vspace{-17pt}
%\vspace{1in}
\caption{
% An illustration of interpreting NeRF representation as the 4D tensor, 
% with its approximation and decomposition and the schematic diagram of uncertainty reduction.
\textbf{\textbf{Left}}: The comparison among the NeRF decomposition methods, including the multi-scale grid~\cite{Li2023DenseRS}, tri-plane~\cite{DBLP:journals/corr/abs-2211-11704}, and factorization~\cite{Chen2022TensoRFTR}. 
\textbf{Right}: Given varied camera poses, the uncertainty distribution is highlighted for the original and the factorized NeRF, with its intensity color bar below. 
}
\label{fig:representation}
\vspace{-10pt}
\end{figure}

% A mainstream direction is to employ RGB-D cameras into the dense visual SLAM algorithms~\cite{newcombe2011kinectfusion,whelan2012kintinuous}. The input RGB-D data is incrementally fused through truncated signed distance functions (TSDFs). 
% There are also neural implicit methods 
% to reconstruct better scenes. 
% Another mainstream direction is to only utilize RGB cameras for dense visual SLAM algorithms. The earlier works directly regress camera poses and depth maps from input images. Later works add multi-view geometric constrains into the prediction, improving the geometric consistency and accuracy of the scene. 
% In contrary to the abovementioned explicit scene representation, DIM-SLAM~\cite{Li2023DenseRS} and NICER-SLAM~\cite{DBLP:journals/corr/abs-2302-03594} utilize neural implicit functions for scene representation, which reconstruct more accurate geometric structures. Our method also utilizes neural implicit functions for dense RGB SLAM algorithm. 
\vspace{-5pt}
\noindent \textbf{Neural Implicit Representations for Visual SLAM.} 
% Our contribution encompasses two primary aspects. Neural Radiance Fields (NeRFs) have revolutionized 3D reconstruction and found applications in various domains, including novel view synthesis, surface reconstruction, and editable scenes and avatars. 
% While this formulation reduces geometric uncertainty, it restricts the learning capability and may propagate errors throughout the entire SLAM process (as observed empirically in Figure ~\ref{fig:init-compare}). Moreover, estimating occupancy using trilinear interpolation on stacked grids of varying resolutions incurs a significant computational cost.
% Neural Radiance Fields (NeRFs) have sparked a revolution for 3D reconstruction and find their applications in various
% fields, 
Neural Radiance Fields (NeRFs) have revolutionized 3D reconstruction and found applications in various domains 
such as novel view synthesis~\cite{mildenhall2020nerf,muller2022instant,verbin2022ref,zhang2020nerf++}, surface reconstruction~\cite{yariv2021volume,oechsle2021unisurf,wang2021neus}, and editable scenes and avatars~\cite{liu2021neural,yang2021learning}. 
%However, these methods are limited to offline 3D reconstruction due to their computationally expensive optimization processes. Recent works have incorporated NeRF into the mapping problem of dense RGB-D SLAM systems. 
However, these methods are often restricted to offline 3D reconstruction due to the computationally expensive optimization processes. Recently, some works~\cite{DBLP:conf/iccv/SucarLOD21,DBLP:conf/cvpr/ZhuPLXBCOP22,DBLP:journals/corr/abs-2211-11704,DBLP:journals/corr/abs-2209-07919,Li2023DenseRS,DBLP:journals/corr/abs-2302-03594,rosinol2022nerf,chung2022orbeez} have incorporated NeRF into the mapping problem of the dense RGB-D SLAM system. 
% iMAP is a pioneering work that efficiently and continuously represents the 3D scene using a single MLP, extending even to unobserved regions. NICE-SLAM employs hierarchical voxel grids and pre-trained decoders to handle larger scenes, while ESLAM suggests using compact feature planes instead of voxel grids for improved processing speed. For a visual comparison between voxel grids and feature planes, please refer to Figure ~\ref{fig:representation}. 
iMAP is a pioneering work that utilizes a single MLP to represent the 3D scene efficiently and continuously, extending even to unobserved regions. NICE-SLAM~\cite{DBLP:conf/cvpr/ZhuPLXBCOP22} employs hierarchical voxel grids and pre-trained decoders to generalize to larger scenes. Furthermore,  ESLAM~\cite{DBLP:journals/corr/abs-2211-11704} proposes to replace the voxel grids with compact feature planes, significantly improving the processing speed. For a visual comparison between voxel grids and feature planes, please refer to~\cref{fig:representation}.
%
%It also demonstrates that signed distance function (SDF) contributes to better reconstruction results, similar with iDF-SLAM~\cite{DBLP:journals/corr/abs-2209-07919}. However, these NeRF-based SLAM systems are limited by RGB-D cameras. 
% NeRF-SLAM and Orbeez-SLAM explicitly integrate NeRF with the visual SLAM system, relying solely on RGB cameras, resulting in a complex system architecture. NICER-SLAM, on the other hand, relies on pre-trained geometric models and does not meet real-time requirements. Recently, DIM-SLAM introduced the first dense RGB SLAM system based solely on neural implicit maps. However, this approach reduces the problem to estimating voxel grid occupancy using a single channel, limiting the exploitation of robust and expressive high-dimensional latent features.
% NeRF-SLAM and Orbeez-SLAM explicitly integrate NeRF with the visual SLAM system, relying solely on RGB cameras, resulting in a complex system architecture. 
For dense RGB SLAM, NeRF-SLAM~\cite{rosinol2022nerf} and Orbeez-SLAM~\cite{chung2022orbeez} explicitly integrate NeRF with the visual SLAM systems, resulting in redundant system architectures. 
%
% by only utilizing RGB cameras, NeRF-SLA and Orbeez-SLAM explicitly integrate NeRF with the visual SLAM system, resulting in cumbersome system architecture.
%
NICER-SLAM~\cite{DBLP:journals/corr/abs-2302-03594} relies on pre-trained geometric models and thus can not meet real-time demand. Recently, DIM-SLAM ~\cite{Li2023DenseRS} introduces the first dense RGB SLAM system entirely based on neural implicit maps. However, the problem has been downgraded into estimating voxel grid occupancy using a single channel without exploiting the robustness and expressiveness of high-dimensional latent features. While this formulation curtails the geometric uncertainty, it limits the learning capability and thus might propagate the error to the entire SLAM process (as empirically observed in~\cref{fig:init-compare}). Additionally, such occupancy estimation requires tri-linear interpolation of stacked grids of many different resolutions, leading to an undesirable computational budget.

\vspace{-5pt}
\section{Neural Field Mapping for SLAM: A Theoretical Analysis}
\label{sec: Methodology}
%
% The general math model for SLAM
\vspace{-5pt}
\subsection{Mapping Uncertainty of Dense RGB SLAM}
\vspace{-5pt}
For the visual SLAM problem, given image frames $\boldsymbol{I}$, it involves estimating the camera's poses $\tilde{{\boldsymbol{P}}}$, the frames' color $\tilde{\boldsymbol{I}}$ and depth $\tilde{\boldsymbol{D}}$ to reconstruct map $m$. 
We assume that $\tilde{{\boldsymbol{P}}}=\{\tilde{\boldsymbol{R}}, \tilde{\boldsymbol{T}}\}$ obtained from trackers entails a normally distributed disturbance $\boldsymbol{n}$, \ie,  $\boldsymbol{P} = \tilde{\boldsymbol{P}}+\boldsymbol{n}$,
$\boldsymbol{n} \sim \mathcal{N}(0,\boldsymbol{Q})$, where $\tilde{\boldsymbol{R}}, \tilde{\boldsymbol{T}}$ are rotation and transition matrix, respectively, and covariance matrix $\boldsymbol{Q}$ denotes the noise variance. 
% $\tilde{{\boldsymbol{P}}}_k=\{\tilde{\boldsymbol{R}_k}, \tilde{\boldsymbol{t}_k}\}$
%
Denote the RGB stream as $\boldsymbol{I}_{1:k}$,
% , which starts from the $1_{th}$ frame to the $k_{th}$ frame. 
%
% The task aims to estimate $\tilde{{\boldsymbol{P}}}$, $\tilde{{\boldsymbol{I}}}$
% with the map $m$ with $\tilde{m}=\{\tilde{D}, \tilde{\boldsymbol{I}}\}$, and $\tilde{m}$ depends on reliable $\tilde{{\boldsymbol{P}}}$. 
% with the map $m$  
the visual SLAM problem can be expressed as estimating a conditional joint probability distribution of $\tilde{{\boldsymbol{P}}}_{1:k}$ and $m$, which can be decomposed into two sub-problems including the localization and mapping~\cite{montemerlo2002fastslam}: 
\setlength\abovedisplayskip{5pt}
\setlength\belowdisplayskip{5pt}
\begin{equation}
\label{eq:full-slam}
\begin{aligned}
P(\tilde{{\boldsymbol{P}}}_{1:k}, m|\boldsymbol{I}_{1:k}) = \underbrace{P(\tilde{{\boldsymbol{P}}}_{1:k}|\boldsymbol{I}_{1:k})}_{localization} \cdot \underbrace{ P(m|\tilde{{\boldsymbol{P}}}_{1:k}, \boldsymbol{I}_{1:k})}_{mapping}. 
\end{aligned}
\end{equation}
Specifically, the localization problem aims to predict $\tilde{{\boldsymbol{P}}}_{1:k}$ given the RGB stream, which is thoroughly discussed in previous literature of SLAM~\cite{dissanayake2001solution,bailey2006simultaneous}. 
We focus on the mapping problem to reconstruct the map incrementally with the observed stream. 
Specifically, it can be divided into \textbf{1) Initialization stage}, the robot needs to predict $\tilde{\boldsymbol{I}}$,  $\tilde{\boldsymbol{D}}$ and estimate $\tilde{{\boldsymbol{P}}}$ when the environment is unknown; \textbf{2) On-the-fly mapping stage},  
upon the system's initialization, the robot estimates the initial camera poses $\tilde{{\boldsymbol{P}}}$ continually updated by the off-the-shelf trackers,~\eg,~\cite{chung2022orbeez} onboard sensor suits.
% the mapping problem construction problem unfolded.
%where $\bold{R}$ is the rotation matrix and $\bold{T}$ is the transition matrix. Denote their variance together as a covariance matrix $\boldsymbol{Q}$,
%
% Therefore, the SLAM map construction problem in \cref{eq:full-slam} can be decomposed into two estimation problems, including localization and mapping, 
% assumes the known robot poses with a small disturbance $\boldsymbol{n}$. 
%
In practice, only partial frame observation is cached in a window of size $w$ to achieve real-time operation.  
Recently, NeRF~\cite{mildenhall2020nerf} has been introduced as a compelling solution to the mapping problem for visual SLAM. 
\textit{However, there is, to date, no theoretical analysis available to examine the mapping uncertainty upon the integration of implicit neural mapping}. 

% We propose to analyze the mapping uncertainty of their integration.
Denote NeRF function as $\mathcal{G}$ and its decoder as $\Phi$. 
The mapping problem of \cref{eq:full-slam} then turns into estimating the map $\tilde{m} =  \Phi(\mathcal{G} (\boldsymbol{r}))$ to maximize the posterior probability (\textit{See suppl. material for details}).  
% given the observed inference $\tilde{m}$, 
%
\setlength\abovedisplayskip{5pt}
\setlength\belowdisplayskip{5pt}
\begin{equation}
\label{eq:max_likelihood}
\begin{aligned}
\tilde{m} = \arg\max_{(\Phi, \mathcal{G}, \tilde{\boldsymbol{r}})} P(\tilde{{\boldsymbol{P}}}_{(k-w):k},\boldsymbol{I}_{(k-w):k}|\Phi(\mathcal{G}(\tilde{\boldsymbol{r}}_{(k-w):k}))), 
\end{aligned}
\end{equation}
where $\tilde{\boldsymbol{r}}(t)=\tilde{\boldsymbol{o}}+t\tilde{\boldsymbol{d}}$ are samples drawn from camera rays originated from the center $\tilde{\boldsymbol{o}}$ with normalized direction $\tilde{\boldsymbol{d}}$, $t$ denotes the ray distance from $\tilde{\boldsymbol{o}}$ at $\tilde{\boldsymbol{r}}$, and $w$ denotes the number of frames in the cached window.  
% as the inputs to the implicit mapping process. 
%
% Denote .
% Recall that a point in the space is modeled as a sample taken along the camera ray of normalized direction $\tilde{\boldsymbol{d}}$. 
%
%N samples $\{t\}_N$
% \begin{equation}
% \label{eq:max_likelihood}
% \begin{aligned}
% &\tilde{m} = \arg\max_{(\Phi, \mathcal{G}, \tilde{\boldsymbol{r}})} P(\tilde{{\boldsymbol{P}}}_{(k-w):k},\boldsymbol{I}_{(k-w):k}|m'), \\
% \end{aligned}
% \end{equation}
%
% Noted that the ray direction is also correlated to the disturbed camera pose $\hat{x}$ and thus expressed as $\hat\vec{d}$ in this formula.
% where xxx. At this stage, 
%
\vspace{-5pt}
\begin{theorem}
\label{theorem: to_mdist}
we interpret the implicit map construction as a maximal posterior distribution in \cref{eq:max_likelihood}, which is equivalent to minimizing its quadratic form. Refer to \textit{Suppl. material} for derivation details. 
\end{theorem}
\setlength\abovedisplayskip{12pt}
\setlength\belowdisplayskip{12pt}
\begin{equation}
\label{eq:pre_md}
\begin{aligned}
\arg\min_{(\Phi, \mathcal{G}, \tilde{\boldsymbol{r}})} (m' - \boldsymbol{I}_{(k-w):k})^T\boldsymbol{Q}^{-1}(m' - \boldsymbol{I}_{(k-w):k}), 
\end{aligned}
\end{equation}

\begin{definition}
\label{def:md}
we formulate the neural field mapping problem as a form of mapping uncertainty, defined as Mahalanobis distance ($d_M$) according to \cref{eq:pre_md},
\end{definition}
\setlength\abovedisplayskip{7pt}
\setlength\belowdisplayskip{9pt}
\begin{equation}
\label{eq:dm}
\begin{aligned}
d_M(m', \boldsymbol{I}) =  (m' - \boldsymbol{I}_{(k-w):k})^T\boldsymbol{Q}^{-1}(m' - \boldsymbol{I}_{(k-w):k}) =||m' - \boldsymbol{I}_{(k-w):k}||_{\boldsymbol{Q}}, 
\end{aligned}
\end{equation}
\cref{eq:dm} shows that the mapping uncertainty is derived from the gap between 
% the ground truth map distribution $m$ 
the frames in the cached window and the estimated map $m'$ (parameterized as rendered RGB images) by reducing $d_M$ regarding these frames. According to \cref{theorem: to_mdist}, 
we aim to reduce the mapping uncertainty by substantiating $(\Phi, \mathcal{G}, \tilde{\boldsymbol{r}})$. In our framework (See ~\cref{sec: win}), we propose the factorized representations for $\Phi$ and $\mathcal{G}$ and sampling strategies for $\tilde{\boldsymbol{r}}$. 
\subsection{Mapping Uncertainty Decomposition}
\label{sec: uncertainty_decomp}
\vspace{-5pt}
% \noindent\textbf{Uncertainty decomposition. }
%[Based on \ref{sec} and \ref{}, we now do what ?]

Denote $i_{th}$ sliding cached window as $\Omega_i=\{\boldsymbol{I}_j\}_{j=1}^w$, the global mapping uncertainty $D_M$ regarding all sampled sequence $\boldsymbol{I}$ can be estimated via Monte Carlo approximation,  
\setlength\abovedisplayskip{3pt}
\setlength\belowdisplayskip{3pt}
\begin{equation}
\label{eq:dm_all}
\begin{aligned}
&
D_M = \int d_M \approx \frac{1}{w} \sum_i d_M(\tilde{\boldsymbol{o}_i}, \tilde{\boldsymbol{d}_i}, \tilde{\boldsymbol{t}_i}, \boldsymbol{I}) \approx 
\frac{1}{w} \sum_i ||\Phi(\mathcal{G}(\tilde{\boldsymbol{o}_i}, \tilde{\boldsymbol{t}_i})), \Omega_i||_{\boldsymbol{Q}},
% \\ &
\end{aligned}
\end{equation}
Note that we follow the common practice of NeRF to randomly sample $\tilde{\boldsymbol{d}}$ \wrt~ the predicted $\tilde{\boldsymbol{o}}$ and obtain the estimated $\tilde{\boldsymbol{r}}$, which correspond to all pixels in $\boldsymbol{I}$. Therefore, it becomes another Monte Carlo estimation to capture the overall distribution of $\tilde{\boldsymbol{o}}$. We assume the uncertainty raised by $\tilde{\boldsymbol{d}}$ is well constrained in this problem since rays are randomly sampled inside $\Omega_i$, and thus omitted from the further formulation. 
We can further divide the uncertainty and reallocate them to the initialization stage $D_{init}$ and on-the-fly mapping stage $D_{fly}$ since they entail slightly different uncertainty levels: 

%Since uncertainty raised by $\tilde{\boldsymbol{d}}$ can be mostly attributed to $\tilde{\boldsymbol{o}}$, 
%we approximate the result with omitted $\tilde{\boldsymbol{d}}$ in \cref{eq:dm_all}. This way, we reduce the uncertainty caused by the rotation matrix $\boldsymbol{R}$. 

% %
% {\setlength\abovedisplayskip{-40pt}
% \setlength\belowdisplayskip{-40pt}
\begin{equation}
\label{eq:dm_all_split}
\begin{aligned}
&
D_M = D_{init} + D_{fly} \approx \frac{1}{w}||\Phi(\mathcal{G}(\tilde{\boldsymbol{o}_0}, \tilde{\boldsymbol{t}_0})), \Omega_0||_{\boldsymbol{Q}} + \frac{1}{w}\sum_{i=1}^I ||\Phi(\mathcal{G}(\tilde{\boldsymbol{o}_i}, \tilde{\boldsymbol{t}_i})), \Omega_i||_{\boldsymbol{Q}},
% \\ &
\end{aligned}
\end{equation}

Specifically, the \noindent\textbf{Initialization stage} is characterized by strong uncertainty regarding mutual reliance among three components, \ie implicit function $\Phi, \mathcal{G}$, trajectories $\tilde{\boldsymbol{o}_0}$, and sampling distance along rays $\tilde{\boldsymbol{t}_0}$). Speedy convergence in initialization is an important indication of SLAM's superiority. For the additional trajectories $\tilde{\boldsymbol{o}_0}$ uncertainty compared with the \noindent\textbf{On-the-fly mapping stage}, a sliding window sampling strategy is introduced in~\cref{sec:sws}. After initialization, the initial pose and the learned implicit representation can easily extend to the entire system, where the estimated camera poses $\tilde{\boldsymbol{o}}$ in the later stage is replaced by poses $\hat{\boldsymbol{o}_i}$ obtained from the off-shelf tracers. Given the pose-level uncertainty is reduced, a robust and efficient representation for $\Phi, \mathcal{G}$ would be proposed to constrain the overall mapping uncertainty given partial observation $\Omega_i$ in~\cref{sec: f-map}.

\vspace{-5pt}

% \end{Proposition}
%
\begin{figure}[t]
\centering
\includegraphics[width=1\textwidth]{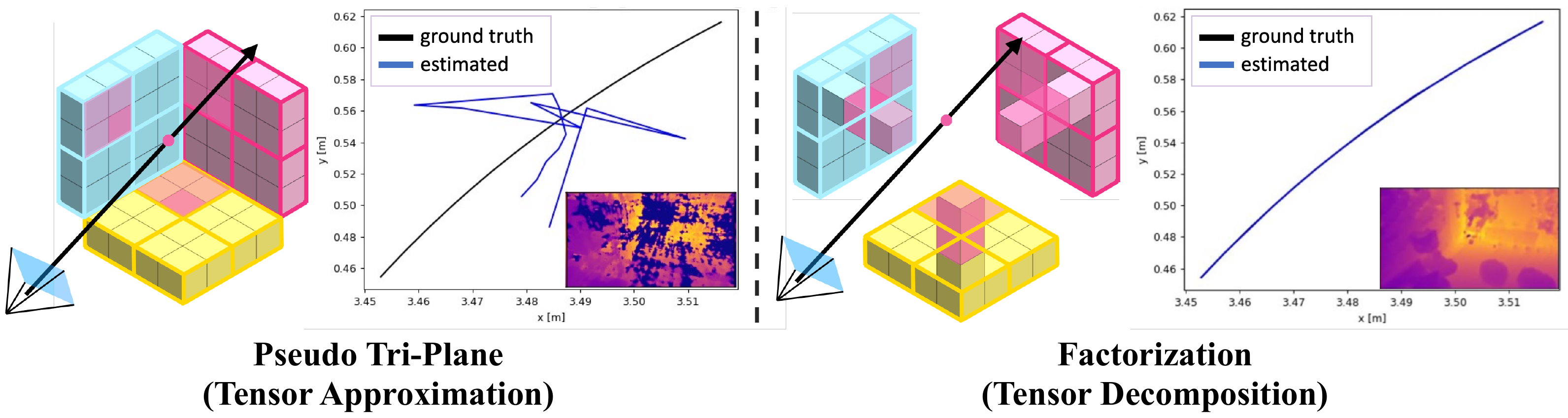}
%\vspace{1in}
\vspace{-17pt}
\caption{Comparison of Tri-projection (\textbf{Left}) and Factorization (\textbf{Right}) in the initialization stage of our FMapping, which is a joint estimation of initial poses and implicit maps. The two resolution layers are adopted by mimicking the experimental setting in ~\cite{DBLP:journals/corr/abs-2211-11704}, the pose trajectory plot and the depth estimation results suggest that the factorized 4D tensors are \textit{much more robust to the uncertainty}.}
\label{fig:observe}
\vspace{-10pt}
\end{figure}

% \vspace{-5pt}
\section{Proposed Framework}
\vspace{-5pt}
\label{sec: method}

\subsection{Factorized Radiance Field}
\label{sec: f-map}
In \cref{sec: uncertainty_decomp}, we explicitly formulate the global mapping uncertainty $D_M$ by \cref{eq:dm_all}.
It turns out that reducing $D_M$ relies on accurate pose estimation $\tilde{\boldsymbol{o}}$ and the robustness of latent feature representation $\mathcal{G}$ against noise. 
In this section, we introduce the factorized neural radiance field as the efficient and reliable 4D representation $\mathcal{G}$ to speed up the computation and relieve the uncertainty in dense RGB mapping.

The first goal is to improve efficiency. It can be achieved by decomposing the NeRF feature computation  
% and 
% handle incomplete observation within the cached window frame.
in $\mathcal{G}$ by multiplying matrix and vector in the lower dimensional space. 
% As shown in , 
%
% we model NeRF implicit function $G$ as a 4D tensor 
%
% approximating it with a smaller number of components,
%
Recent works~\cite{eg3d021,Chen2022TensoRFTR,DBLP:journals/corr/abs-2211-11704} leverage the
matrix decomposition to speed up NeRF computation, \ie,~ 
% as the 3D space sample distribution with its latent feature dimension 
representing the high-dimensional features by samples' 3D coordinates along with their latent feature, 
which has emerged as a prevailing technique for NeRF acceleration. 
It has been proven that factorization is more efficient, which decomposes the tri-linear interpolation into the bi-linear and linear~\cite{DBLP:journals/corr/abs-2211-11704} (also see \cref{fig:representation}), while preserving its expressiveness. 
%
%
% In ~\cref{fig:observe}, we empirically evaluate the two common decomposition paradigms, namely the Tri-plane representation that projects the 3D tensor onto three 2D feature planes; and the BTD-based Factorized representation. The experimental results show that the Factorization scheme is extremely robust during the initialization stage, and thus would forms a sufficiently good representation in our entire mapping process for dense RGB SLAM.
%
Therefore, we leverage factorization to enhance computational efficiency in NeRF computation without dampening the rendering fidelity. 

Secondly, we aim to reduce the uncertainty when the pose $\tilde{\textbf{o}}$ is unreliable.  
Previously, multi-resolution grids~\cite{DBLP:conf/cvpr/ZhuPLXBCOP22, Li2023DenseRS} (See \cref{fig:representation}) is used to constrain the uncertainty in SLAM. 
Unfortunately, such representation highly depends on geometry priors, either from external sensors and pre-trained models or strict constraints on spatial coherence, which is vulnerable to noise.
To address it, we apply factorization to lessen the noise by passing the uncertainty into the lower-dimensional space.
To verify it, we compare the factorization against the bi-linear used in Tri-plane with the experiment shown in \cref{fig:observe}. 
It turns out that the factorization is more robust to high uncertainty in the map initialization stage. 
We refer readers to the \textit{suppl. material.} for further analysis of its robustness to noise.

In practice, given NeRF function $\mathcal{G}$ and its decoder $\Phi$, jointly denoted as parameters $\theta$, density and colors are estimated by underlying continuous volumetric scene function $\sigma_\theta(\tilde{\boldsymbol{x}})$ and $c_\theta(\tilde{\boldsymbol{x}}, \tilde{\boldsymbol{d}})$. 
Volume rendering~\cite{mildenhall2020nerf} aims to constrain the spatial coherence by projecting the estimated samples $\tilde{\boldsymbol{x}}$ along rays $\tilde{\boldsymbol{r}}$ for color supervision, 
\begin{equation}
\label{eq:volrend}
\begin{aligned}
& \tilde{\boldsymbol{I}}(\tilde{\boldsymbol{r}})= \sum_{i=1}^{N}\alpha_\theta(\tilde{\boldsymbol{x}_i})\prod_{j < i}\left(1-\alpha_{\theta}(\tilde{\mathbf{x}_{j}})\right)c_\theta(\tilde{\mathbf{x}_{i}},\tilde{\boldsymbol{d}}), \\
& \alpha_{\theta}(\tilde{\boldsymbol{x}_i}) = 1-exp(-\sigma_{\theta}(\tilde{\boldsymbol{x}_i})\delta_i), 
\end{aligned}
\end{equation}
where $\alpha_\theta(\tilde{\boldsymbol{x}_i})$ denotes the penetrating light at $\tilde{\boldsymbol{x}_i}$. 
% of $\tilde{\boldsymbol{x}_i}$ sampled from arbitrary ray $\tilde{\boldsymbol{r}}$. 
% (t)=\tilde{\boldsymbol{o}}+t\tilde{\boldsymbol{d}}$ with estimated camera parameters $\tilde{\boldsymbol{o}} = \{\tilde{\boldsymbol{o}_k}\}$,  
% $\tilde{\boldsymbol{d}} = \{\tilde{\boldsymbol{d}_{ij,k}}\}$
% in \cref{eq:cam_ray}.  
%
Then $\alpha(\boldsymbol{x})
$ composites the predictions into the rendered frames, supervised by the corresponding color frames.
Therefore, the rendering loss constraints the spatial coherence by rendering each view correctly $\mathcal{L}_c = \sum_k ||\tilde{\boldsymbol{I}_k} - \boldsymbol{I}_k||_2^2$, 
However, the cross-frame information is not fully leveraged using $\mathcal{L}_c$ since it doesn't constrain the uncertainty for spatial coherence within the cached window frames.
% only ensuring each view is rendered correctly.
A recent work~\cite{Li2023DenseRS} proposes to enforce geometric consistency using a wrapping loss $\mathcal{L}_w$ with the spirit of multi-view stereo~\cite{6909592}, \ie,  
lifting the 2D pixel in frames into 3D space and then projecting it to another frame. \textit{We refer readers to Suppl. material for further derivation and analysis}. 
The total loss is denoted as $\mathcal{L} =\beta_{c}\mathcal{L}_c +\beta_{w}\mathcal{L}_w$, where $\beta$ represents a weighting factor. 

To evaluate depth and color, by following ~\cite{DBLP:conf/cvpr/ZhuPLXBCOP22, DBLP:journals/corr/abs-2302-03594}, $\tilde{\boldsymbol{D}_k}$ and  $\tilde{\boldsymbol{I}_k}$ \wrt~\cref{eq:volrend} are rendered as: 
\begin{equation}
\label{eq:depth_color}
\begin{aligned}
\tilde{\boldsymbol{D}_k} = \sum_{i=1}^{N}w_\theta(\tilde{\boldsymbol{x}_i})t_i, \ 
\tilde{\boldsymbol{I}_k} = \sum_{i=1}^{N}w_\theta(\tilde{\boldsymbol{x}_i})c_\theta(\tilde{\mathbf{x}_{i}},\tilde{\boldsymbol{d}}),
\end{aligned}
\end{equation}
where $w_\theta(\tilde{\boldsymbol{x}_i})=\alpha_\theta(\tilde{\boldsymbol{x}_i})\prod_{j < i}\left(1-\alpha_{\theta}(\tilde{\mathbf{x}_{j}})\right)$. 
% Please refer to \cref{prop:ray_level uncertainty} for further discussion on termination probability\cite{oechsle2021unisurf}. 
$\alpha_\theta(\tilde{\boldsymbol{x}_i})$(\cref{eq:volrend}) entails uncertainty regarding sample distribution $\delta$ which relies on inverse transform sampling (\cref{eq:pdf_sampling}). Thus, \cite{oechsle2021unisurf} uses the sigmoid activation directly on $\sigma_\theta(\tilde{\boldsymbol{x}_i})$ to avoid the ambiguity, \ie, 
$\alpha_\theta(\tilde{\boldsymbol{x}_i})$ is replaced by $O_\theta(\tilde{\boldsymbol{x}_i}) = Sigmoid(\sigma_\theta(\tilde{\boldsymbol{x}_i}))$ in \cref{eq:depth_color}.
\textit{We refer interested readers to Suppl. material for further analysis on termination probability $O_\theta(\tilde{\boldsymbol{x}_i})$}. 
Notably, different from RGB-D SLAM with direct depth supervision $||\boldsymbol{D}-\tilde{\boldsymbol{D}}||_1$ ~\cite{DBLP:conf/iccv/SucarLOD21, DBLP:conf/cvpr/ZhuPLXBCOP22, DBLP:journals/corr/abs-2211-11704}, our problem lacks the geometry cues and relies solely on RGB inputs.

% Then, the radiance field tensor $\mathcal{G}$ with the component size $R$ can be decomposed into low-rank tensor components through the VM decomposition, 
 
%\begin{equation}
%\begin{aligned}
%& \mathcal{G}(x)= \sum_{r=1}^R %(\boldsymbol{v}_r^X\circ\boldsymbol{M}_r^{YZ}+\boldsymbol{v}_r^Y\circ\boldsymbol{M}_r^{XZ}+\boldsymb%ol{v}_r^Z\circ\boldsymbol{M}_r^{XY})\circ\boldsymbol{b}_r,
%\end{aligned}
%\end{equation}
%
%where the rank-one tensor components are denoted as $\boldsymbol{v}_r^{X}\in \mathbb{R}^I$, %$\boldsymbol{v}_r^{Y}\in \mathbb{R}^J$, $\boldsymbol{v}_r^{Z}\in \mathbb{R}^K$; %$\boldsymbol{M}_r^{YZ}\in \mathbb{R}^{J\times K}$, $\boldsymbol{M}_r^{XZ}\in \mathbb{R}^{I\times K}$,
%$\boldsymbol{M}_r^{XY}\in \mathbb{R}^{I\times J}$ are matrix factors; $\boldsymbol{b}_r \in \mathbb{R}^{1+P}$ is the corresponding latent feature, $\circ$ represents the outer product, and is an arbitrary . 

\label{sec: win}

\vspace{-5pt}
\subsection{Sliding Window Sampling}
\vspace{-5pt}
\label{sec:sws}

\begin{figure}[t]
\centering
\includegraphics[width=1\textwidth]{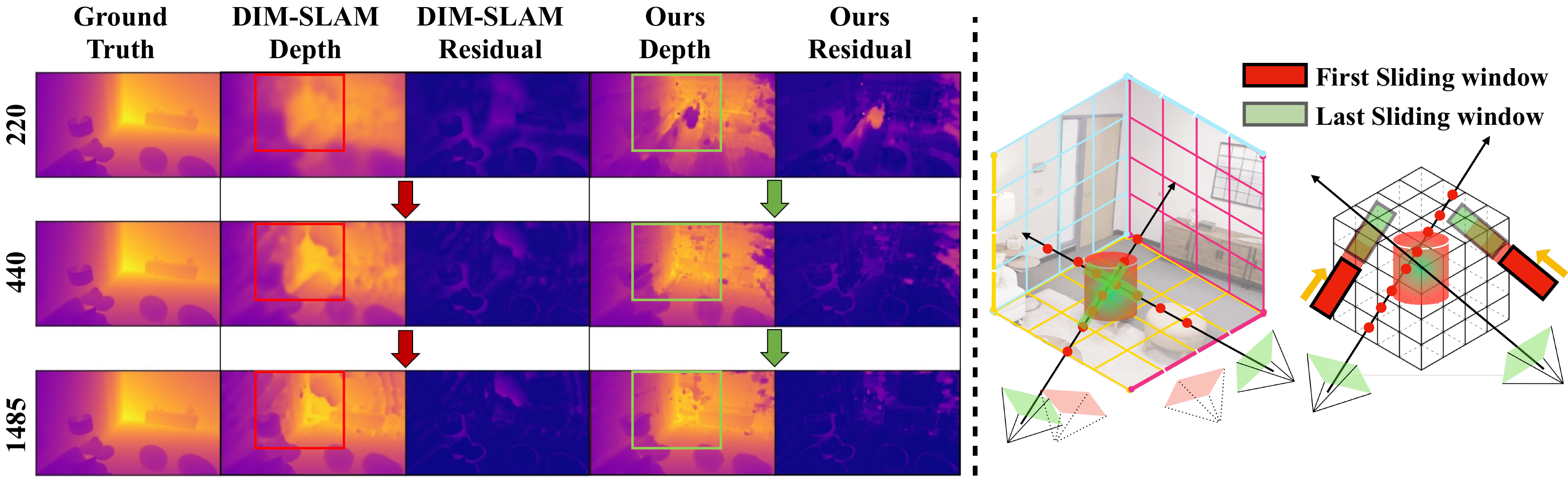}
%\vspace{1in}
\caption{\textbf{Left:} Depth estimation in the initialization stage and the schematic plot of sliding window sampler. Results are plotted for the 220, 440, and 1485 iterations for DIM-SLAM and our methods during system initialization, where only the supervision signal is color. DIM-SLAM is reinforcing the inaccurate geometric estimation due to insufficient feature channels, while FMapping shows a stronger learning capability to rule out the artifacts.
\textbf{Right:} Sliding window sampler is represented as a red window kernel that moves along the ray directions and marches on until detected discontinuity.}

%Denote the $\tilde{\boldsymbol{o}_m}$ as the green camera centers and $\tilde{\boldsymbol{o}_{m+1}}$ as the red ones. Given the bias in the pose estimation between the training iteration $m$ and $m+1$, the sliding window sampler explores the coherent geometric cues, \ie, the shared object intervals within the highlighted green areas. Specifically, the sliding window is represented as a window kernel as the red rectangle that moves along the estimated ray directions. }

\label{fig:init-compare}
\end{figure}

During the rendering process, $N$ samples $X_r=\{\tilde{\boldsymbol{x}_i}|\tilde{\boldsymbol{x}_i}=\tilde{\boldsymbol{r}}(t_i), t_i < t_{i+1}\}_{i=1}^N$ along any estimated ray $\tilde{\boldsymbol{r}}$ are drawn from the coarse stratified sampling, followed by the inverse transform sampling according to the coarse-level sampling $\mathcal{F}_{pdf}(\tilde{\boldsymbol{x}})$ over its normalized PDF $\alpha_\theta(\tilde{\boldsymbol{x}})$,  
\begin{equation}
\label{eq:pdf_sampling}
\begin{aligned}
X_{r, k+1} = \mathcal{F}_{pdf}^{-1}(u) \cup X_{r, k}, \  
\mathcal{F}_{pdf}(\tilde{\boldsymbol{x}_{i, k+1}}) = \sum_{i} P(\tilde{\boldsymbol{x}_{i, k}}|\alpha_\theta(\tilde{\boldsymbol{x}_{i, k}})<u), u\in \boldsymbol{U}, 
\end{aligned}
\end{equation}
where $\boldsymbol{U} \sim Unif[0, 1]$, and $k$ denotes the iteration times for multi-stage estimation, \eg, $k=2$ in the coarse-to-fine hierarchical sampling\cite{mildenhall2020nerf}. The resulting adjacent sample distance in $X_r$ as $\delta_i = |\boldsymbol{x}_{i+1}-\boldsymbol{x}_{i}|$ is shown in \cref{eq:volrend}. 
$X_{r, k}$ is sorted after each sampling iteration. 

As we discussed in \cref{sec: uncertainty_decomp}, the map initialization stage faces strong uncertainty, $\mathcal{F}_{pdf}$ is not ideal regarding its limited ability to handle the accumulated error during optimization, since both pose estimation and implicit representation are learned from scratch. 
The sliding sampler aims to \textit{reduce the mapping uncertainty in this stage}. 
% pose-level(\cref{prop:pose_level uncertainty}) and ray-level(\cref{prop:ray_level uncertainty}) 
In detail, 
the proposed sampler smooths the noise brought by uncertainty and exploits the continuity of object intervals through the iterative training process. 
Specifically, we utilize the sliding window kernel to reduce the noise. Then, we explore the potential objects' boundary intervals by searching the densely distributed samples with rendering contributions, attributed to the estimated camera poses and updated $F$ among the training iterations. 
% 
%
% Notably, the sliding window sampling takes advantage of evaluated 
%
% For sampling, 
%
% 1. DIM-SLAM->NeuS: multiple-stage pdf sampling
%
% 2. Unisurf: occlusion activated from density 
Specifically, 
as illustrated in \cref{fig:init-compare}, the iterative sampling process consists of a stratified sampling $\mathcal{F}_{sw}$ with estimated interval range $\boldsymbol{t}_{r, k}$, along with the inverse transform sampling (\cref{eq:pdf_sampling}), 
\begin{equation}
\label{eq:sliding_window_sampling}
\begin{aligned}
& X_{r, k+1} = \mathcal{F}_{pdf}^{-1}(\boldsymbol{U}) \cup \mathcal{F}_{sw}(\boldsymbol{t}_{r, k}) \cup X_{r, k}, \\ &  
\mathcal{F}_{sw}(\boldsymbol{t}_{r, k}) = \tilde{\boldsymbol{o}}+(\boldsymbol{t}_{r, k, n} + u (\boldsymbol{t}_{r, k, f}-\boldsymbol{t}_{r, k,  n}))\tilde{\boldsymbol{d}}, u\sim Unif(0, 1), 
\end{aligned}
\end{equation}
%
% where $u$ is a random number. 
Moreover, the sliding window searches for high-reward intervals on the smoothed scores which are distributed densely in the space. 
Denote a window kernel $\boldsymbol{G}=\tau\times[\frac{1}{\tau}]$ with size $\tau$. $X_{r, k}$ (\cref{eq:pdf_sampling}), 
% Is sorted based camera distances $t_i$, 
the sliding window $SW$ is defined as, 
\begin{equation}
\label{eq:sliding_window_operation}
\begin{aligned}
& SW(X_{r, k}) = \boldsymbol{W} \star \boldsymbol{G} = \sum_i \boldsymbol{W}_{i, r, k}\cdot \boldsymbol{G}_{\tau-i}, \\
& \boldsymbol{W}_{r, k+1} = \beta * \exp(-\boldsymbol{\delta}_k) +(1-\beta) * w_\theta(X_{r, k}), \\ 
\end{aligned}
\end{equation}
where $\boldsymbol{\delta}_k$ is the adjacent sample distances from the last ray sample $X_{r, k}$, and $\star$ is the convolution operation. We can derive $\boldsymbol{t}_{r, k}$ by densely searching within qualified regions, enriched by filtered scores ${\boldsymbol{W}}$ (see \textit{suppl. material} for technical details).
%is defined by searching for the filtered scores $\hat{\boldsymbol{W}_{r, k+1}}$ that is continued distributed in the space, 
and $\beta$ is a hyper-parameter to balance the contribution of the rendering distance and weight components.
For further comparison between the sampling strategies, please refer to \textit{suppl. material}. 
% \noindent\textbf{Residual guidance. }

% \textbf{\textcolor{blue}{To quantify the uncertainty:
% \\
% 1. parameter $\theta$: exact values are unknown to experimentalists and cannot be controlled. 
% \\
% 2. inverse sampling: mutual reliance $\theta$ and $\boldsymbol{x}$. 
% \\
% 3. Forgetting problem \& local window with on-the-fly camera: MLP 
% }}

% \textbf{\textcolor{red}{(FastSLAM notes)May be shifted into Appendix}}

% 1. Define SLAM as a probabilistic Markov chain
% motion model: pose evolve

% \textbf{\textcolor{red}{Our solution}}
 
% xxx, 

\subsection{System Overview}
\label{sec:overview}
 % \subsection{Volume Rendering}

%  \begin{figure}[t]
% \centering
% \includegraphics[width=1\textwidth]{framework.png}
% %\vspace{1in}
% \caption{The working pipeline of FMapping.}
% \label{fig:init-compare}
% \end{figure}

\textbf{The initialization stage:} We collect 15 frames for jointly estimating initial poses and local implicit maps. At this stage, 4 iterations of sampling are performed, where the stratified samples are collected at the first iteration for rough distribution estimation, followed by 3 iterations of sliding window sampling. The sample sliding window size is set to 3, 4, and 5 for the second, third, and last iteration, respectively. The sampling sizes of each stage are 32, 64, 64, and 64, respectively. For each sliding window sampling iteration, $40\%$ samples are first retrieved from the estimated distribution, then the remaining $60\%$ are smoothly picked through the sliding window along the rays. Once the initialization stage is finished, the first frame is added to the global keyframe set and kept fixed. In addition, the parameters of the MLP decoder are frozen after initialization, similar to ~\cite{Li2023DenseRS}. 
 
% K(1-4)
%1/ stratified 
%2-4 40pdf 60 sw (2[3],3[4],4[5])  3阶段采样需要采样（ 1-3 ）【32，64，64，64】
%经过sw进行

\textbf{The on-the-fly mapping stage: } In the process of mapping, we maintain an active window of 20 frames, with the portion of the global and local frame the same with ~\cite{Li2023DenseRS}. 20 iterations of optimization are performed to update the map for every 5 frames. The oldest 5 local frames are removed while the new 5 incoming frames are added to the window for the next map update.

\vspace{-8pt}
\section{Experiments}
\vspace{-8pt}

We comprehensively evaluate our method on simulated dataset~\cite{straub2019replica}, and real-world dataset~\cite{sturm2012benchmark}. For all datasets, we quantitatively and qualitatively compare our FMapping to the state-of-the-art NeRF-based SLAM methods, \ie, iMAP~\cite{DBLP:conf/iccv/SucarLOD21} and NICE-SLAM~\cite{DBLP:conf/cvpr/ZhuPLXBCOP22} that were re-implemented under our experimental settings. \textit{For the comparisons in no-pose and no-depth scenarios, please refer to the suppl. material for details.}  

% \textbf{\textcolor{red}{To be added:  }}

\subsection{Experimental Setting}

\noindent \textbf{Dataset and Evaluation Metrics.} We evaluate our method on Replica dataset~\cite{straub2019replica} and TUM RGB-D dataset~\cite{sturm2012benchmark}. We choose to evaluate the same scenes as done in DIM-SLAM~\cite{Li2023DenseRS}. We utilize three 3D scene reconstruction metrics: reconstruction accuracy [cm], reconstruction completion [cm], and completion ratio [< 5 cm \%], following the previous work~\cite{DBLP:conf/cvpr/ZhuPLXBCOP22}. \textit{Due to page limitation, qualitative results on the TUM RGB-D dataset can be found in the suppl. material.}
%(w). We split these datasets followed NICE-SLAM~\cite{DBLP:conf/cvpr/ZhuPLXBCOP22}.

% \textbf{\textcolor{red}{To zidong: please update the tables by 8 am}}

\noindent \textbf{Implementation Details.} All experiments are conducted on a single NVIDIA RTX 3090 GPU. The factorized representation is inspired and implemented based on the TensoRF~\cite{Chen2022TensoRFTR}. The resolution of the feature grid is computed based on the defined bounding box. We implement a single-resolution factorized feature grid with each dimension set to 128. To make it comparable with existing neural implicit mapping methods using a voxel grid, the resolution is roughly calculated as a voxel size of $\sim$ 8cm, given a bounding box of size 11.8m, 8.7m, and 6.8m for three coordinates (the example is given for Replica scene room 0). The feature channels are set to 16 and 24 for density and appearance components, respectively. The color decoder is a two-layer MLP with 48 channels in the hidden layer. Adam optimizer~\cite{kingma2014adam} is adopted with learning rates set to 0.02 for the grid feature updating and set to 0.001 for color decoder updating, respectively.

\subsection{Evaluation of Mapping}

As shown in \cref{fig:Comparison}, NICE-SLAM demonstrates strong reliance on the availability of depth supervision and thus fails in our experimental re-implementation. Therefore, we only compare the scene reconstructions with re-implemented iMAP, which is the directly comparable neural field mapping paradigm with ours, since both frameworks could serve as the only scene representation for the dense SLAM problem. 
\begin{table}[t!] 
\footnotesize
  \caption{Quantitative comparison of our proposed method's mapping performance on Replica indoor scenes with re-implemented iMAP(-) given the same experimental setting (\ie no depth supervision available for mapping provided given accurate pose).}
  \label{recons}
  \setlength{\tabcolsep}{1.2mm}
  \centering
  \begin{tabular}{cc|ccccccc|c}
    \toprule
    Method & & room1 & room2 & office0 & office1 & office2 & office3 & office4 & Avg.
    \\
    \midrule
    \multirow{3}{*}{iMAP(-)} &  Acc. $\downarrow$ & 9.28 & 7.36 & 5.27 & 7.37 & 7.09 & 6.26 & 6.84 & 7.07
    \\
    &  Comp. $\downarrow$ & 7.80 & 8.47 & 6.13 & 7.68 & 6.83 & 7.73 & 7.83 & 7.50
    \\
    &  Comp. Ratio (\%) $\uparrow$ & 45.94 & 43.39 & 56.00 & 41.58 & 51.30 & 49.04 & 50.98 & 48.32
    \\
    % \midrule
    % \multirow{3}{*}{NICE-SLAM~\cite{DBLP:conf/cvpr/ZhuPLXBCOP22}} &  \textbf{Scene Acc.} [cm] $\downarrow$
    % \\
    % &  \textbf{Scene Comp.} [cm] $\downarrow$
    % \\
    % &  \textbf{Scene Comp. Ratio} [<5cm \%]  $\uparrow$ 
    % \\
    \midrule
    \multirow{3}{*}{FMapping (Ours)} &  Acc. $\downarrow$ & 23.02 & 19.92 & 15.75 & 14.10 & 15.06 & 16.22 & 13.11 & 16.74
    \\
    &  Comp. $\downarrow$ & \textbf{2.94} & \textbf{4.28} & \textbf{3.38} & \textbf{3.18} & \textbf{3.58} & \textbf{4.33} & \textbf{4.11} & \textbf{3.69}
    \\
    &  Comp. Ratio (\%) $\uparrow$ & \textbf{90.60} & \textbf{71.54} & \textbf{82.07} & \textbf{83.45} & \textbf{81.73} & \textbf{73.97 }& \textbf{75.23} & \textbf{79.80}
    \\
    \bottomrule
  \end{tabular}
  \vspace{-5pt}
\end{table}
 \begin{figure}[t!]
\centering
\includegraphics[width=1\textwidth]{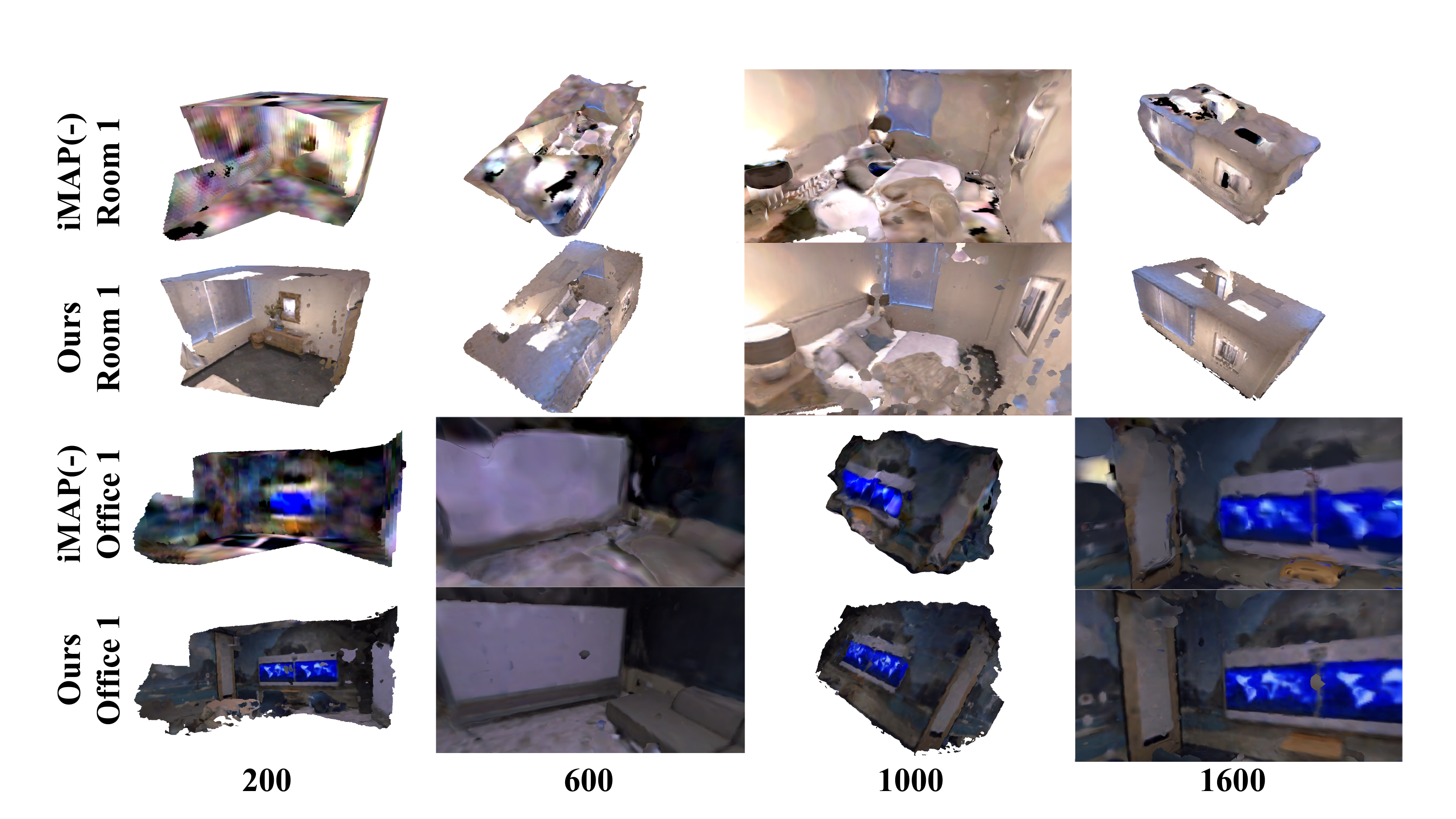}
%\vspace{1in}
\caption{Visualization of dense RGB mapping snapshots (at 200, 600, 1000, and 1600 input frames) of the on-the-fly running iMAP (-) and our FMapping in Replica (Room1 and) sequences, given
ground truth poses without depth supervision.}
\label{fig:acc}
\end{figure}

FMapping achieves an overall highest completion ratio, as shown in \cref{recons}, which is almost 2 times better than iMAP(-). Notably, as a dense colorized point cloud mapping, our mesh reconstruction accuracy is of the same order of magnitude compared to iMAP(-), with better reconstruction visualization throughout the entire SLAM process, providing a real-time high-fidelity perception, as shown in \cref{fig:acc}. 

We also evaluated the performance of our FMapping with iMAP(-) on the condition that both depth supervision and estimated pose are not available, where the $\{\boldsymbol{R}, \boldsymbol{T}\}$ are optimized solely relying on the back-propagation of color gradient. The experimental results show that FMapping is still able to perform reconstruction of significantly higher fidelity compared to iMAP(-), which fails to reconstruct necessary geometry. \textit{See Suppl. material for more details}.

% \begin{table}[h] 
% \footnotesize
%   \caption{Main table.}
%   \label{main-table}
%   \setlength{\tabcolsep}{1.5mm}
%   \centering
%   \begin{tabular}{ccccccc}
%     \toprule
%     & $\text{TSDF-Fusion}^{*}$ & iMAP & $\text{iMAP}^{*}$ & NICE-SLAM & $\text{NICE-SLAM}^{*}$ & fMAP\\
%     \midrule
%     \textbf{Scene Acc.} [cm] $\downarrow$ & 1.28 & 4.43 & 2.15 & 2.94 & 3.04 &  \\
%     \textbf{Scene Comp.} [cm] $\downarrow$ & 5.61 & 5.56 & 2.88 & 4.02 & 3.84 &  \\
%     \textbf{Scene Comp. Ratio} [<5cm \%]  $\uparrow$ & 82.67 & 79.06 & 90.85 & 86.73 & 86.52 &  \\
%     \bottomrule
% \end{tabular}
% \end{table}

% \begin{table}[h] 
% \footnotesize
%   \caption{Memory, computation, and running time.}
%   \label{test-table}
%   \setlength{\tabcolsep}{1.5mm}
%   \centering
%   \begin{tabular}{cccc}
%     \toprule
%     Method & Memory(MB)$\downarrow$ & FLOPs($\times 10^3$) & Mapping(ms)$\downarrow$\\
%     \midrule
%     iMAP~\cite{DBLP:conf/iccv/SucarLOD21} & 1.02 & 443.91 & 448(1000, 10, 44)  \\
%     NICE-SLAM~\cite{DBLP:conf/cvpr/ZhuPLXBCOP22} & 12.0(16cm) & 104.16 & 130(1000, 10, 48)  \\
%     DIM-SLAM~\cite{DBLP:journals/corr/abs-2302-03594} (two threads) & 9.1(16cm) & 18.76 & 330(3000, 100, 96)  \\
%     DIM-SLAM~\cite{DBLP:journals/corr/abs-2302-03594} (one thread) & 29.1(8cm) & 18.76 & 335(3000, 100, 96)  \\
%     Ours & & 90.72 & 2010(3000, 100, 224) \\
%     \bottomrule
% \end{tabular}
% \end{table}
%
\subsection{Evaluation of Speed, Computation, and Memory}
In \cref{tab:sample-table}, we first evaluate the mapping speed between our method and state-of-the-art works including iMAP~\cite{DBLP:conf/iccv/SucarLOD21}, NICE-SLAM~\cite{DBLP:conf/cvpr/ZhuPLXBCOP22}, and DIM-SLAM~\cite{Li2023DenseRS}. Our FMapping can process online mapping with 20 frames in 2.01 seconds, which is comparable with iMAP~\cite{DBLP:conf/iccv/SucarLOD21}. Note that the mapping problem of our RGB SLAM has more uncertainty than RGB-D SLAM, and thus we need more sampling points during training. Then, we compare the FLOPs for computing on one 3D point. Our method requires only 1/5 FLOPs of iMAP, which is more efficient. We ascribe it to the factorized neural field representation. Finally, regarding memory consumption, our method is about $\times1000$ slighter than NICE-SLAM and DIM-SLAM, and about $\times100$ than iMAP, making it available to scenarios with limited computational resources. Note that we achieve real-time performance with the generalized Pytorch implementation, while DIM-SLAM requires tailored CUDA acceleration. \textit{Due to the space limit, we refer readers to the suppl. material for detailed theoretical analysis, framework, and experimental results.} 
 
% performance of our mapping method and compare it with state-of-the-art works including iMAP~\cite{DBLP:conf/iccv/SucarLOD21}, NICE-SLAM~\cite{DBLP:conf/cvpr/ZhuPLXBCOP22}, and DIM-SLAM~\cite{Li2023DenseRS}. 
%
% Regarding memory consumption, our method is about $1000\times$ slighter than NICE-SLAM and DIM-SLAM, and about $100\times$ than iMAP, making it available to scenarios with limited computational resources. 
%
% Furthermore, FMapping can process online mapping with 10 seconds per frame, which is around $5\times$ faster than NICE-SLAM, and around $2\times$ than iMAP. 
% Despite DIM-SLAM has lower computation FLOPs than our approach, we surpass it by a large margin on parameter size. 
%
%

\begin{table}[t] 
\setlength{\tabcolsep}{15pt}
\footnotesize
  \caption{Analysis of our method in comparison with existing ones in terms of mapping speed, FLOPs, number of parameters, and model size growth rate (parameterized by scene side-length $L$).}
  \label{tab:sample-table}
  \centering
  \begin{tabular}{ccc|cc}
    \toprule
    \multirow{2}{*}{Method} & Mapping speed & Computation  & \multicolumn{2}{c}{Memory}\\
    & s / frame num. & FLOPs ($\times 10^3$) & \# Param. & Grow. R. \\
    \midrule
    iMAP~\cite{DBLP:conf/iccv/SucarLOD21} & 0.48/5 & 443.91 & 0.22 M & - \\
    NICE-SLAM~\cite{DBLP:conf/cvpr/ZhuPLXBCOP22} & 0.13/5 & 104.16 & 12.18 M & $O(L^3)$ \\
    DIM-SLAM~\cite{Li2023DenseRS} & - & 18.76 & 29.10 M & $O(L^3)$  \\
    Ours & 2.01/20 & 80.64 & \textbf{0.01 M} & $O(L^2)$ \\
    \bottomrule
\end{tabular}
\end{table}

\vspace{-8pt}
\section{Conclusion and Future Work}
\vspace{-5pt}
In this paper, we presented FMapping, a light and efficient neural field mapping for real-time dense RGB SLAM. We leveraged the factorized neural field representation to decompose uncertainty into a lower-dimensional space, enhancing robustness to noise and improving training efficiency. A sliding sampler ensured a robust initialization, which enabled the initialization of map reconstruction using coherent geometric cues, thus allowing a continuous robust mapping throughout the entire SLAM process. We achieved high-fidelity mapping with $\times 20$ and $\times 1000$ fewer parameters than the current implicit neural mapping design for RGB-D and RGB input streams, respectively.
% In this paper, we explore dense SLAM with only the RGB camera. We focused on the mapping problem and build up a theoretical analysis of the mapping uncertainty. 
% Based on it, we then propose the factorized neural field with a sliding window strategy to reduce the uncertainty for scene reconstruction. 
% In the future, we plan to take three parts including initialization, tracking, and mapping in dense RGB SLAM into consideration. 
% %
% We hope our contributions from both theoretical and technical aspects could shed some light on this field when encountering problems about how to reduce the mapping uncertainty, improve the system efficiency, as well as using consumer-level devices (\eg, RGB cameras).
In this paper, we investigated dense SLAM using only an RGB camera. Specifically, we focused on the mapping problem and provided a theoretical analysis of mapping uncertainty. Building upon this analysis, we proposed a factorized neural field combined with a sliding window strategy to mitigate uncertainty during scene reconstruction. 
%

%\textbf{Limitation \& Future Works:} 
%In this work, we only focus on the mapping of the complete dense RGB SLAM problem consisting of three components: initialization, tracking, and mapping. In this regard, our method may not be effective enough to address the uncertainty of camera poses brought by the offline tracers when the camera is on the fly.  

%In the future, 
%we plan to consider the whole dense RGB SLAM process and propose a solution with the uncertainty analysis regarding the system. 

%\textbf{Broader impact:} We hope our contributions, encompassing both theoretical and technical aspects can provide valuable insights and guidance in addressing challenges related to reducing implicit mapping uncertainty and enhancing system efficiency for dense RGB SLAM.

% \section{Limitation and Future Work}

\clearpage
{
\small
\bibliographystyle{plain}
\bibliography{egbib}
}

\end{document}